\documentclass[10pt,twocolumn,letterpaper]{article}

 \usepackage{iccv}              

\definecolor{iccvblue}{rgb}{0.21,0.49,0.74}
\usepackage[pagebackref,breaklinks,colorlinks,allcolors=iccvblue]{hyperref}
\usepackage{stfloats}
\usepackage{multirow} 
\def\paperID{*****} 
\def\confName{ICCV}
\def\confYear{2025}

\title{CT-GRAPH: Hierarchical Graph Attention Network for Anatomy-Guided \\ CT Report Generation}

\author{
Hamza Kalisch$^{1,2}$ \quad
Fabian Hörst$^{1,3}$ \quad
Jens Kleesiek$^{1,3}$ \\
Ken Herrmann$^{2}$ \quad
Constantin Seibold$^{1}$ \\
$^{1}$Institute for AI in Medicine (IKIM), University Hospital Essen (AöR), Essen, Germany \\
$^{2}$Department of Nuclear Medicine, University Hospital Essen (AöR), Essen, Germany \\
$^{3}$Department of Physics, TU Dortmund, Dortmund, Germany \\
{\tt\small \{hamza.kalisch, fabian.hoerst, jens.kleesiek, ken.herrmann\}@uk-essen.de} \\
{\tt\small constantin.seibold@gmail.com}
}

\begin{document}
\maketitle
\begin{abstract}
As medical imaging is central to diagnostic processes, automating the generation of radiology reports has become increasingly relevant to assist radiologists with their heavy workloads. Most current methods rely solely on global image features, failing to capture fine-grained organ relationships crucial for accurate reporting. To this end, we propose CT-GRAPH, a hierarchical graph attention network that explicitly models radiological knowledge by structuring anatomical regions into a graph, linking fine-grained organ features to coarser anatomical systems and a global patient context. Our method leverages pretrained 3D medical feature encoders to obtain global and organ-level features by utilizing anatomical masks. These features are further refined within the graph and then integrated into a large language model to generate detailed medical reports. We evaluate our approach for the task of report generation on the large-scale chest CT dataset CT-RATE. We provide an in-depth analysis of pretrained feature encoders for CT report generation and show that our method achieves a substantial improvement of absolute 7.9\% in F1 score over current state-of-the-art methods. The code is publicly available at \href{https://github.com/hakal104/CT-GRAPH}{https://github.com/hakal104/CT-GRAPH}.
\end{abstract}    
\section{Introduction}
\label{sec:intro}
Radiology reports play a pivotal role in clinical workflows, translating complex imaging findings into actionable diagnostic insights for clinical personal. However, manual documentation of 3D computed tomography (CT) scans takes up considerable amounts of time \cite{goergen2013evidence}. While automated methods have advanced significantly for X-rays \cite{liu2019clinically,huang2023kiut}, extending these successes to 3D CTs remains challenging.

One key challenge is the high computational cost of processing 3D volumes. Most existing approaches \cite{wu2023towards, hamamci2024ct2rep, chen2024dia}  rely on end-to-end training pipelines that are memory-intensive and require long training times. To reduce these demands, some methods fall back on workarounds such as operating on significantly downsampled volumes \cite{wu2023towards, chen2024dia} or processing 2D slices rather than full 3D volumes \cite{lee2024read}. However, the former discards anatomical details because of its reduced resolution, whereas the latter compromises spatial coherence and overlooks long-range anatomical relationships that are crucial for holistic analysis.

Another key limitation of many existing methods \cite{bai2024m3d, wu2023towards, hamamci2024ct2rep, chen2024dia, liu2024benchmarking} is their sole reliance on global representations, which are passed directly to the language model for report generation. This neglects fine-grained anatomical features that are critical for capturing organ-specific details, thereby limiting the model’s ability to localize findings and reason about distinct regions within the scan. Our experiments show that integrating such fine-grained features consistently improves performance over global representations alone. Notably, we observe substantial gains on localized pathologies such as consolidation and pericardial effusion, where spatial precision and region-aware context are essential for accurate clinical reporting. 
Region-level features have been used in X-Ray report generation, first via bounding boxes \cite{tanida2023interactive} and more recently through segmentation masks \cite{gu2024orid}. Similarly, for CT report generation, recent methods such as Reg2RG \cite{chen2024large} and VividMed \cite{luo2024vividmed} incorporate region-wise information to improve fine-grained feature extraction. In contrast to Reg2RG, our method not only models the relationships between regional features and the global context, but also effectively utilizes anatomical masks at finer granularity (e.g., lung lobes instead of entire lungs), enabling more precise and localized feature representations.

This highlights the need for more efficient and anatomically informed methods for CT report generation that can better capture both local detail and global context. A promising solution is a two-stage approach that first leverages frozen visual features to reduce computational cost. These features can be obtained, for example, from pretrained 3D volumetric encoders and subsequently used to guide report generation. This setup significantly lowers training time and memory requirements while still capturing rich spatial information. To this extent, we explore the potential of pretrained 3D feature encoders to capture fine-grained organ-level features.

However, while such a two-stage approach improves efficiency, it alone is insufficient for capturing the complex anatomical structures and their dependencies. To address this, we propose a hierarchy-aware framework that integrates these pretrained features into anatomically structured graph representations. By explicitly modeling both local detail and global anatomical context, our graph-based method structures features into an anatomical hierarchy, leading to improved clinical accuracy. The main contributions of this work are summarized in the following:

\begin{enumerate}
    \item We leverage frozen, pretrained 3D feature encoders for both global and organ-specific feature extraction via 3D mask pooling, preserving fine-grained spatial information.
    \item We conduct an extensive study on the effective use of pretrained features for multi-abnormality classification and CT report generation. 
    \item We propose CT-GRAPH, a graph attention network (GAT) that hierarchically aggregates fine-grained organ features (e.g., lung lobes) into coarse anatomical regions (e.g., lungs), enabling region-guided report generation. Our experiments on the CT-RATE \cite{hamamci2024developing} dataset show that the proposed hierarchical graph method consistently outperforms competitive baselines, achieving an F1 improvement of 7.9\% over state-of-the-art approaches for CT report generation.
\end{enumerate}
\section{Related work}
\label{sec:formatting}

\paragraph{CT report generation.}
 
As the first approach to 3D CT report generation, CT2REP \cite{hamamci2024ct2rep} leverages an auto-regressive causal transformer for 3D visual extraction and a memory-driven decoder to generate detailed reports from global volume features. Dia-LLaMA \cite{chen2024dia} incorporates diagnostic prompts as medical priors, enabling symptom-driven reporting.  Similarly, AG-RG \cite{di2024ct} proposes an abnormality-guided approach where the report is generated based on predicted abnormalities. In M3D \cite{bai2024m3d} the patch tokens of the Vision Transformer are first reduced with a 3-D average-pooling layer, after which an MLP maps them to the LLM’s embedding size.
By contrast, RadFM \cite{wu2023towards} feeds the tokens into a Perceiver module that compresses them into 64 learnable query vectors. Most recently, Argus~\cite{liu2024benchmarking}  investigates the effectiveness of different pretraining and visual token compression strategies. They introduce a family of vision-language models specifically tailored to 3D report generation, that are able to efficiently process high-resolution 3D images.
A common limitation across these methods is their exclusive use of global image representations, which overlooks organ-specific details essential for anatomy-aware report generation. In contrast, our approach explicitly incorporates fine-grained anatomical features to address this gap.

\paragraph{Region-based methods.} Prior work in X-ray report generation has explored region-level features, initially through the use of bounding boxes~\cite{tanida2023interactive} and more recently via segmentation masks to provide more precise anatomical guidance~\cite{gu2024orid, gu2024complex, seibold2022detailed}. Specifically, ORID \cite{gu2024orid} introduces an organ-region-driven framework for radiology report generation that integrates cross-modal organ-level information using a importance weighting module to suppress noise from unrelated regions and improve diagnostic relevance.
Recent methods for CT report generation, including Reg2RG~\cite{chen2024large} and VividMed~\cite{luo2024vividmed}, utilize region-level cues to refine feature representations and capture more detailed anatomical information. VividMed~\cite{luo2024vividmed} enables region-level visual grounding by supporting both semantic segmentation and instance-level bounding boxes across 2D and 3D modalities, using a three-stage training pipeline to enhance report generation. Reg2RG~\cite{chen2024large} introduces a region-guided framework for CT report generation that leverages anatomical masks to extract local features from referring regions via cropping. These features are combined with global representations to improve contextual understanding and support report generation through a region-report alignment strategy. While Reg2RG \cite{chen2024large} incorporates regional information, it does not explicitly model fine-grained anatomical structures or their relationships to the global context in a structured way. In contrast, our approach hierarchically organizes regional features and integrates them with the global context in a more structured and coherent manner.

\paragraph{Self-supervised 3D vision encoders.} Recent advances in self-supervised learning \cite{haghighi2021transferable, tang2022self, wu2024voco, goncharov2023vox2vec} have enabled the pretraining of 3D vision encoders on large-scale medical datasets without requiring manual annotations. Methods such as VoCo \cite{wu2024voco, wu2024large} have demonstrated strong generalization across various downstream tasks. These encoders capture volumetric context and structural priors, making them well-suited for complex medical analysis. Frozen features from such pretrained models offer a computationally efficient alternative to end-to-end training, while retaining rich anatomical representations essential for downstream tasks like report generation. We leverage these features in a multi-stage training procedure by first extracting globally- as well as anatomically-pooled features for our graph generation.

\section{Methodology}
\label{sec:methodology}

\begin{figure*}[t]
\centering
\includegraphics[width=\textwidth]{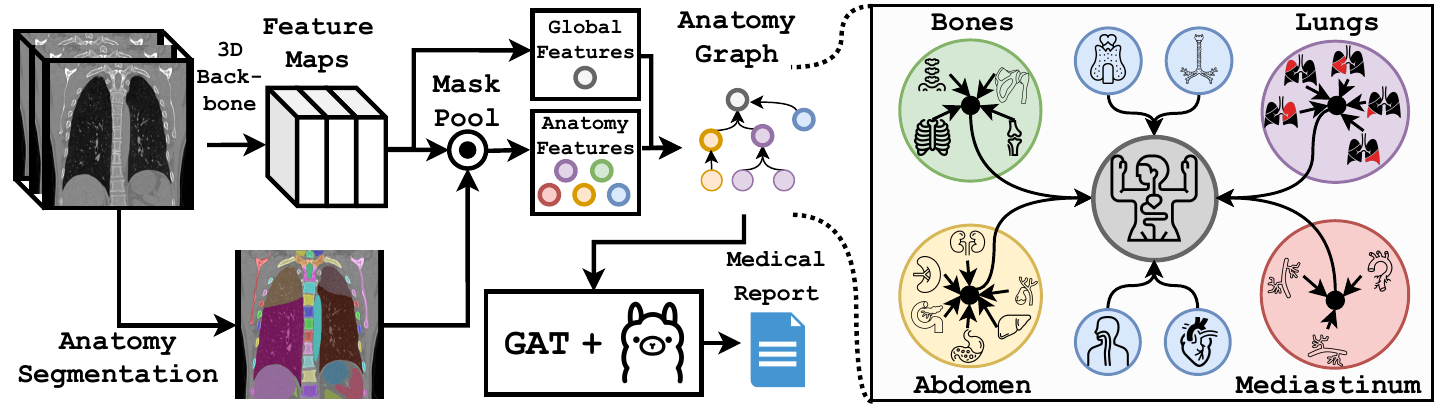}
\caption{Illustration of our main method. The left visualizes the report generation pipeline. We extract 3D feature maps using a generic frozen encoder, and anatomical masks using TotalSegmentator. From these, we obtain regional features via mask-based pooling and a global feature via volume-level aggregation. These features form the basis of a hierarchical anatomical graph, which is processed by a GAT and then passed to a LLaMA model to generate the medical report. On the right, the hierarchical graph structure is shown, with organs (e.g. spleen, liver...) being connected towards corresponding broader anatomical regions (e.g. abdomen), which subsequently connect to a global node. } \label{method_fig}
\end{figure*}

Our framework (see Fig. \ref{method_fig}) first extracts both global and local features from the input CT volume using a frozen, pretrained 3D encoder. The global representation is obtained by aggregating the full volume into a compact feature embedding, while local, organ-specific features are extracted through mask-based pooling over anatomical regions of interest. The extracted features are then organized into a hierarchical graph, where nodes represent anatomical regions at different levels of granularity: fine-level structures (e.g., lung lobes), coarse-level groupings (e.g., lungs), and a global context node. Edges are defined by a fixed anatomical hierarchy, linking each fine-level node to its corresponding coarse-level parent and ultimately to the global node. A graph attention network (GAT) is applied to this graph to propagate contextual information across different anatomical levels. Finally, the graph features are combined with an embedded textual input prompt and passed to a large language model (LLM) which generates a radiology report.

\subsection{Feature Extraction}
\label{sec:features}

Given a 3D CT volume \( X \in \mathbb{R}^{H \times W \times D} \) of height \( H \), width \( W \), and depth \( D \), and its multi-label segmentation mask \( M \in \{0,1,\ldots,K\}^{H \times W \times D} \), where \( K \) is the number of anatomical structures, our method extracts features via three stages.

\paragraph{Global Feature Encoding.}  
A pretrained 3D feature encoder processes \( X \) to produce hierarchical feature maps \( F^l \) at resolution \( (H_l, W_l, D_l) \) and channel dimension \( C_l \), where \( H_l, W_l, D_l \) denote the height, width, and depth of the 3D feature map at layer \( l \), respectively. The global feature map is defined as the output feature from the final layer \( L \).

\paragraph{Organ Mask Pooling.}  
For each anatomical structure \( k \in \{1,\ldots,K\} \), we resize the multi-label mask \( M \) to the encoder’s spatial resolution of layer $l$ via nearest-neighbor interpolation, obtaining \( M_l \) of size \( H_l \times W_l \times D_l \). The organ-specific feature \( f_k^l \in \mathbb{R}^{C_l} \) is computed via mask-guided average pooling:  
\begin{equation}
f_k^l = \frac{1}{|\mathcal{V}_k^l|} \sum_{(i,j,t) \in \mathcal{V}_k^l} F^l[i,j,t,:],
\end{equation}
where \( \mathcal{V}_k^l = \{(i,j,t) \mid M_l[i,j,t] = k\} \) denotes the voxels of organ \( k \) at resolution level \( l \), and \( |\mathcal{V}_k^l| \) is the organ’s voxel count.

\paragraph{Layer Fusion.}  
Mask pooling as displayed in Eq. (1) is applied at every layer to produce layer-specific features \( f_k^l \in \mathbb{R}^{C_l} \). These multi-scale features are concatenated along the channel dimension to form the final organ representation \( f_k = [f_k^1; \ldots; f_k^L] \in \mathbb{R}^{C_{total}} \), combining fine-grained textures from early layers with semantic context from deeper layers. \( C_{total} \) is defined by \( \sum_{l=1}^L C_l \).

\subsection{Hierarchical Graph Attention Network}
\label{subsec:hierarchical_gat}

Radiological findings occur at different anatomical levels, ranging from localized abnormalities (e.g., lung nodule) to organ-level pathologies (e.g., atelectasis). To capture this hierarchy, our graph model aggregates information bottom-up: fine-grained features (e.g., upper left lung lobe) contribute to higher-level representations (e.g., the whole lung), enabling a structured integration of local abnormalities into a broader diagnostic context.

\paragraph{Graph Structure and Node Definitions.}
Our graph architecture organizes anatomical structures into three hierarchical levels (see Fig. 1):

\begin{itemize}
    \item \textbf{Fine-level Nodes} ($\mathcal{F}$): These nodes represent individual anatomical structures, with each fine-level node corresponding to an anatomical mask. The feature pooled from this region is processed by an multi-layer perceptron (MLP) to produce the corresponding node embedding $h_f$.
    \item \textbf{Coarse-level Nodes} ($\mathcal{C}$): Coarse-level nodes represent broader anatomical systems or regions (e.g., the abdomen). The corresponding anatomical mask is formed by the union of the masks of fine-level nodes belonging to the same system. A feature is extracted via mask pooling from these combined masks and passed through an MLP to obtain the coarse-level embedding $h_c$. Some anatomical structures (e.g., the esophagus) do not belong to any specific system; thus, we use them as coarse-level node and link them directly to the global node. 
    \item \textbf{Global Node} ($v_g$): The global node represents the entire scan. Its feature map is derived from a downsampled version of the global feature map, obtained by applying adaptive average pooling to reduce the map to a size of (4, 4, 2). These pooled features are then passed through a multi-layer perceptron (MLP) to obtain the final feature embedding for the global node.
\end{itemize}

\paragraph{Hierarchical Connectivity.}  
Edges \( \mathcal{E}_{fc} \) and \( \mathcal{E}_{cg} \) in the graph represent hierarchical anatomical relationships. Fine-level nodes are connected to their corresponding coarse-level nodes, and coarse-level nodes are connected to the global node:

\begin{equation}
\mathcal{E}_{fc} = 
\underbrace{\{(v_f, v_c) \mid v_c \in \mathcal{C},\, v_f \in \mathcal{F}_{v_c} \}}_{\text{fine} \rightarrow \text{coarse}}, 
\label{eq:efc}
\end{equation}

\begin{equation}
\mathcal{E}_{cg} = 
\underbrace{\{(v_c, v_g) \mid v_c \in \mathcal{C}\}}_{\text{coarse} \rightarrow \text{global}}.
\label{eq:ecg}
\end{equation}

Here, $\mathcal{F}_{v_c}$ denotes the subset of all fine-level nodes $v_f$ assigned to coarse node $v_c$. The full edge set of the hierarchical graph is given by the union of \( \mathcal{E}_{fc} \) and \( \mathcal{E}_{cg} \).

\paragraph{Graph Attention Mechanism.}

Message passing occurs through a graph attention mechanism \cite{velivckovic2017graph}, applied hierarchically between fine-level nodes, coarse-level nodes and the global node.

In the fine-to-coarse attention step, each coarse node \( v_c \) computes an updated feature $h_c'$ by aggregating the features of its corresponding fine-level nodes \( v_f \in \mathcal{F}_{v_c} \). Self-loops are included to allow each node to attend to itself:

\begin{equation}
h_c' = \sum_{v_f \in \mathcal{F}_{v_c} \cup \{v_c\}} \alpha_{v_f v_c} \, W h_f.
\label{eq:f2c-update}
\end{equation}
The attention coefficients \( \alpha_{v_f v_c} \) are computed as

\begin{equation}
\begin{split}
\alpha_{v_f v_c} =
\frac{\displaystyle
      \exp\!\bigl(\text{LReLU}(a^\top [W h_f \parallel W h_c])\bigr)}
     {\displaystyle
      \sum_{v \in \mathcal{F}_{v_c} \cup \{v_c\}}
      \exp\!\bigl(\text{LReLU}(a^\top [W h_v \parallel W h_c])\bigr)},
\end{split}
\label{eq:attn-coeff}
\end{equation}

where \( W \) is a learnable weight matrix, LReLU refers to Leaky ReLU \cite{xu2015empirical}, \( a \) is a shared attention vector, and \( \parallel \) denotes feature concatenation. Layer Normalization \cite{ba2016layer} is applied to all node features before applying attention.

The coarse-to-global attention is applied analogously, with the global node updated by aggregating the features of the coarse nodes. A skip connection is added to preserve the original global features. We employ multi-head attention and concatenate the resulting outputs across heads for both the fine-to-coarse and coarse-to-global aggregation.

\paragraph{Report Generation.} The output graph features from the global, coarse-level, and fine-level nodes are first projected into the LLM’s latent space. These features are then concatenated with the embedded input prompt \( p \), which is defined as: "Generate a medical report based on the visual information of the given CT image.". The resulting combined input \( x = [g; p] \) is subsequently fed into the LLM, which is trained to minimize cross-entropy loss for next-token prediction in the context of anatomy-guided CT report generation.

\section{Preliminaries}
\label{sec:experiments}

\subsection{Datasets and preprocessing}

CT-RATE \cite{hamamci2024developing} is a publicly available, large-scale dataset consisting of 3D chest CT volumes paired with corresponding medical reports. It includes 25,701 non-contrast CT volumes from 21,314 unique patients, expanding to 49,138 volumes through multiple reconstructions optimized for various window settings. We retain only CT volumes with non-duplicate and non-empty reports. For these selected volumes, we use reports from the Radgenome-chest CT \cite{zhang2024radgenome} dataset, which also provides region-wise reports for the exact same CT scans. In total, our training set consists of 22,778 samples, with a test set of 1,505 samples. CT-RATE also offers abnormality labels, which categorizes each CT volume into multiple of 18 different pathologies. We further split the training data into 90\% for training and 10\% for validation.

\subsection{Evaluation protocol}

For pathology classification, we assess the performance of pretrained features on the multi-label classification task using linear probing. Evaluation metrics include macro-averaged F1 score, precision, and recall, which treat each class equally and highlight performance across both common and rare pathologies. 
For report generation, we utilize clinical efficacy (CE) metrics, including macro-averaged F1, precision, and recall computed on labels extracted from generated reports. Label extraction is performed using the pathology classifier introduced in CT-CLIP \cite{hamamci2024developing}. In addition, we report natural language generation (NLG) metrics such as BLEU-1 to BLEU-4 \cite{papineni2002bleu}, METEOR \cite{banerjee2005meteor}, and ROUGE-L \cite{lin2004rouge}. These evaluate lexical similarity, semantic alignment, and sentence-level structure compared to reference reports. Reported values are evaluated on the test set.

\subsection{Implementation details}

We resize all volumes to a size of 512x512x256 using trilinear interpolation and apply further normalization depending on the pretrained feature encoder being used.
For multi-scale feature extraction, pretrained features are obtained via sliding window inference with a patch overlap of 0.25. We utilize TotalSegmentator \cite{wasserthal2023totalsegmentator} to obtain anatomical segmentation masks from the CT volumes. For our hierarchical graph structure we define the bones, lungs, abdomen, mediastinum, heart, esophagus, trachea, and thyroid as the coarse nodes. The detailed listing of the 34 fine-grained nodes can be found in the code. 
For pathology classification, we employ linear probing with a batch size of 32, training for 40 epochs with a learning rate of 0.0001 and weight decay of 0.0001 to mitigate overfitting. For report generation, we use the base LLaMA2-7B \cite{touvron2023llama} model as our language backbone and apply LoRA \cite{hu2022lora} for efficient fine-tuning. In addition to LoRA adapters, we also allow the embedding layer and output head to remain trainable. We use a LoRA rank of 32, a scaling factor of 32, and a dropout rate of 0.1. The model is optimized using AdamW with a fixed learning rate of $5 \times 10^{-5}$. Training for 6 epochs with pre-extracted features and a batch size of 8 takes approximately 10 hours on a single NVIDIA A100 GPU.

\begin{table}[t]
    \centering
    \scriptsize
    \caption{Comparison of pre-trained 3D feature encoders.}
    \label{tab:encoders}
    \renewcommand{\arraystretch}{1.1}
    \setlength{\tabcolsep}{5pt}
    \begin{tabular}{lccccc}
        \toprule
        Feature Encoder & Architecture & Dataset Size & Feature Dimensions \\
        \midrule
        SwinUNETR \cite{tang2022self} & SwinViT \cite{liu2021swin} & 5,050 & [48,96,192,384,768] \\
        VoCo-10k \cite{wu2024voco} & SwinViT \cite{liu2021swin} & 10,500 & [48,96,192,384,768]  \\
        VoCo-160k \cite{wu2024large} & SwinViT \cite{liu2021swin} & 160,167 & [48,96,192,384,768] \\
        Vox2Vec \cite{goncharov2023vox2vec} & FPN \cite{goncharov2023vox2vec} & 6,550 & [16,32,64,128,256,512] \\
        TransVW \cite{haghighi2021transferable} & 3D U-Net \cite{ellis2020trialing} & 623 & [64,128,256,512]  \\
        CT-FM \cite{pai2025vision} & SegResNet \cite{myronenko20183d} & 148,000 & [32,64,128,256,512]  \\
        \bottomrule
    \end{tabular}
\end{table}
\section{Experimental Results}
\label{sec:results}

 \begin{table*}[t]
    \centering
    \scriptsize
    \caption{Performance comparison of feature encoders for pathology classification across different layers and layer fusion.}
    \label{tab:layers}
    \renewcommand{\arraystretch}{1.1}
    \setlength{\tabcolsep}{4pt}
    \begin{tabular}{l ccccccc}
        \toprule
        \multirow{2}{*}{\textbf{Feature Encoder}}  
        & \multicolumn{6}{c}{\textbf{Layers}} & \multirow{2}{*}{\textbf{Layer Fusion}} \\
        \cmidrule(lr){2-7}
        & Layer 1 & Layer 2 & Layer 3 & Layer 4 & Layer 5 & Layer 6 & \\
        \midrule
        SwinUnetr\cite{tang2022self} & 0.097 & 0.219  & 0.235  & 0.251 & 0.243 & / &  \textbf{0.350} \\
        VoCo-10k\cite{wu2024voco} & 0.118 & 0.249 & 0.305 & 0.289 & 0.171 & / & \textbf{0.394} \\
        VoCo-160k\cite{wu2024voco} & 0.118 & 0.245 & 0.306 & 0.297 & 0.185 & / & \textbf{0.394} \\
        Vox2Vec \cite{goncharov2023vox2vec} & 0.303 & 0.331 & 0.358 & 0.367 & 0.387 & 0.350 & \textbf{0.441} \\
        TransVW \cite{haghighi2021transferable} & 0.172 & 0.217 & 0.273 & 0.351 & /  & / & \textbf{0.372} \\
        CT-FM \cite{haghighi2021transferable} & 0.051 & 0.173 & 0.275 & 0.397 & 0.429  & / & \textbf{0.448} \\
        \midrule
    \end{tabular}
\end{table*}
 \begin{table}[b]
    \centering
    \scriptsize
    \caption{Performance comparison of different feature encoders across pooling levels.}
    \label{tab:feat_evaluation}
    \renewcommand{\arraystretch}{1.1}
    \setlength{\tabcolsep}{4pt}
    \begin{tabular}{l l ccc}
        \toprule
        \multirow{2}{*}{\textbf{Feature Encoder}} & \multirow{2}{*}{\textbf{Pooling Level}} &
        \multicolumn{3}{c}{\textbf{CE Metrics}} \\
        \cmidrule(lr){3-5}
        & & P & R & F1 \\
        \midrule
        \multirow{3}{*}{SwinUnetr\cite{tang2022self}}   
            & Global  & 0.309 & 0.011 & 0.020 \\
            & Coarse  & \textbf{0.718} & 0.155 & 0.226 \\
            & Fine    & 0.604 & \textbf{0.259} & \textbf{0.341} \\
        \midrule
        \multirow{3}{*}{VoCo-10k\cite{wu2024voco}}    
            & Global  & 0.281 & 0.019 & 0.010 \\
            & Coarse  & \textbf{0.728} & 0.167 & 0.238 \\
            & Fine    & 0.687 & \textbf{0.287} & \textbf{0.375} \\
        \midrule
        \multirow{3}{*}{VoCo-160k\cite{wu2024voco}}    
            & Global  & 0.274 & 0.010 & 0.019 \\
            & Coarse  & 0.631 & 0.176 & 0.247 \\
            & Fine    & \textbf{0.679} & \textbf{0.301} & \textbf{0.385} \\
        \midrule
        \multirow{3}{*}{TransVW\cite{haghighi2021transferable}}   
            & Global  & 0.000 & 0.000 & 0.000 \\
            & Coarse  & \textbf{0.717} & 0.203 & 0.278 \\
            & Fine    & 0.668 & \textbf{0.315} & \textbf{0.371} \\
        \midrule
        \multirow{3}{*}{Vox2Vec\cite{goncharov2023vox2vec}}    
            & Global  & 0.293 & 0.036 & 0.060 \\
            & Coarse  & \textbf{0.592} & 0.279 & 0.345 \\
            & Fine    & 0.558 & \textbf{0.384} & \textbf{0.434} \\
        \midrule
        \multirow{3}{*}{CT-FM\cite{pai2025vision}}    
            & Global  & 0.507 & 0.074 & 0.116  \\
            & Coarse  & \textbf{0.613} & 0.286 & 0.372  \\
            & Fine    & 0.496 & \textbf{0.386} & \textbf{0.431} \\
        \bottomrule
    \end{tabular}
\end{table}

\subsection{Feature Aggregation}

The effectiveness of our hierarchical graph depends critically on the representational quality of the input features used to model anatomical regions. To this end, we evaluate different strategies to extract features from pretrained encoders by linear probing on the multi-label pathology classification task.  We consider three pooling strategies: (1) Global pooling, applying global average pooling over the entire feature map; (2) Coarse-level pooling, using mask pooling with coarse anatomical masks; and (3) Fine-level pooling, using mask pooling with fine-level masks. 

We first investigate the impact of encoder layer depth by extracting pooled features from different individual layers of the network. For each layer, the total mask-pooled features from all pooling-levels are utilized as input for classification. 
We subsequently evaluate the effect of pooling granularity by comparing global, coarse, and fine-level pooling using multi-scale fused features as input for pathology classification. 
As pretrained feature encoders we consider six options: SwinUNETR \cite{tang2022self}, VoCo \cite{wu2024voco, wu2024large} (with 10k and 160k pretraining samples respectively), Vox2Vec \cite{goncharov2023vox2vec}, TransVW \cite{haghighi2021transferable}, and CT-FM \cite{pai2025vision}. Details regarding architecture, pretraining dataset size and feature dimensions of different layers can be seen in Table \ref{tab:encoders}.

\paragraph{Effect of Layers.} The results (see Table \ref{tab:layers}) indicate that higher layers generally yield better performance for pathology classification, with features from the penultimate and last layer producing the strongest results. Applying Layer Fusion consistently improves performance across all feature encoders, with particularly notable gains for architectures such as VoCo and SwinUNETR. Interestingly, VoCo-10k and VoCo-160k achieve nearly identical performance despite a large gap in pretraining dataset size. Vox2Vec shows strong performance even in earlier layers and CT-FM achieves the largest F1 score overall.

\paragraph{Effect of Pooling Level.}

Tab. \ref{tab:feat_evaluation} shows that fine-level features exhibit the strongest performance for Recall and F1, indiciating its effectiveness in capturing localized pathological patterns. Except for VoCo-160k, coarse-level pooling yields the highest precision across all feature encoders. In contrast, global pooling leads to drastically reduced performance, with near-zero recall and F1 scores, confirming that excessive spatial compression discards critical anatomical information.

\subsection{Ablations on the Graph Construction}

To investigate the impact of structured input representations on report generation, we conduct a series of ablation studies that vary in how anatomical and spatial information is encoded and integrated into the LLM. Each variant produces a different set of input features, which are then passed to the language model. In addition to our full method (CT-GRAPH), we evaluate the following baselines:

\begin{itemize}
    \item \textbf{Global}: A single token is obtained by compressing the global feature map via a 2-layer MLP. This serves as the standard global representation used in all ablations that include a global token, unless otherwise specified.
    
    \item \textbf{Global (multiple)}: All spatial tokens from the global feature map are projected to the LLM input dimension via a 2-layer MLP.
    
    \item \textbf{Local}: All mask-pooled region features are projected via a 2-layer MLP and used as input tokens to the LLM.
    
    \item \textbf{Global + Local}: Global and local features are separately projected via independent MLPs, and the resulting tokens are concatenated before being fed into the LLM.
    
    \item \textbf{Random Graph}: A graph is constructed with randomly assigned edges between fine-grained nodes, removing any anatomical structure.
    
    \item \textbf{Single-Level Hierarchy Graph}: A simplified graph where all fine-level nodes connect directly to the global node, omitting the intermediate coarse-level hierarchy.

    \item \textbf{CT-GRAPH}: Our proposed graph-based method that hierarchically aggregates fine-grained region features to coarse-level nodes and then to the global context using graph attention. 

\end{itemize}

\begin{table*}[t]
    \centering
    \scriptsize
    \caption{Performance comparison of different training setups leading up to our proposed method (CT-GRAPH) using VoCo-160k features. }
    \label{tab:ablations}
    \renewcommand{\arraystretch}{1.1}
    \setlength{\tabcolsep}{4pt}
    \begin{tabular}{lcccccc|ccc}
        \toprule
        \multirow{3}{*}{\textbf{Method}} & \multicolumn{6}{c|}{\textbf{NLG Metrics}} & \multicolumn{3}{c}{\textbf{CE Metrics}} \\
        \cmidrule(lr){2-7} \cmidrule(lr){8-10}
        & BLEU-1 & BLEU-2 & BLEU-3 & BLEU-4 & ROUGE-L & METEOR & Precision & Recall & F1 \\
        \toprule
        Global (compressed) & 0.4660 & 0.3610 & 0.2903 & 0.2370 & 0.3073 & 0.4088 & 0.282 & 0.133  & 0.171 \\
        Global (multiple) & 0.4582 & 0.3551 & 0.2855 & 0.2333 & 0.3044 & 0.4014 & 0.227 & 0.081  & 0.111 \\
        Local & 0.4685 & 0.3624 & 0.2904 & 0.2365 & 0.3053 & 0.4084 & 0.338 & 0.182 & 0.230 \\
        Global + Local & 0.4751 & 0.3702 & 0.2986 & 0.2446 & 0.3089 & 0.4159 & 0.343 & 0.201 & 0.245 \\
        Random Graph & 0.4690 & 0.3631 & 0.2918 & 0.2384 & 0.3103 & 0.4092 & 0.328 & 0.172 & 0.216 \\
        Single-Level hierarchy graph & 0.4743 & 0.3651 & 0.2914 & 0.2360 & 0.3060 & 0.4097 & 0.385 & 0.202 & 0.253 \\
        CT-GRAPH & \textbf{0.4850} & \textbf{0.3765} & \textbf{0.3024} & \textbf{0.2467} & \textbf{0.3126} & \textbf{0.4205} & \textbf{0.396} & \textbf{0.248} & \textbf{0.296} \\
        \bottomrule
    \end{tabular}
\end{table*}
\begin{table*}[b]
    \centering
    \scriptsize
    \caption{Performance comparison for report generation of different feature encoders using our proposed pipeline.}
    \label{tab:feat_evaluation_rg}
    \renewcommand{\arraystretch}{1.1}
    \setlength{\tabcolsep}{4pt}
    \begin{tabular}{lcccccc|ccc}
        \toprule
        \multirow{2}{*}{\textbf{Feature Encoder}}  
        & \multicolumn{6}{c|}{\textbf{NLG Metrics}} & \multicolumn{3}{c}{\textbf{CE Metrics}} \\
        \cmidrule(lr){2-7} \cmidrule(lr){8-10}
        &  BLEU-1 & BLEU-2 & BLEU-3 & BLEU-4 & ROUGE-L & METEOR & Precision & Recall & F1  \\
        \midrule
        SwinUNETR\cite{tang2022self} & 0.4730 & 0.3656 & 0.2931 & 0.2390 & 0.3077 & 0.4106 & 0.359 & 0.200 & 0.249 \\
        VoCo-10k\cite{wu2024voco} & 0.4779 & 0.3687 & 0.2950 & 0.2398 & 0.3077 & 0.413 & 0.394 & 0.212 & 0.264 \\
        VoCo-160k\cite{wu2024voco} & \textbf{0.4850} & \textbf{0.3765} & \textbf{0.3024} & \textbf{0.2467} & \textbf{0.3126} & \textbf{0.4205} & 0.396 & \textbf{0.248} & \textbf{0.296} \\
        Vox2Vec\cite{goncharov2023vox2vec} & 0.4707 & 0.3654 & 0.2943 & 0.2408 & 0.3090 & 0.4102 & 0.364 & 0.187 & 0.238\\
        TransVW\cite{haghighi2021transferable} & 0.4779 & 0.3685 & 0.2943 & 0.2383 & 0.3039 & 0.4132 & \textbf{0.421} & 0.236 & 0.292 \\
        CT-FM\cite{pai2025vision} & 0.4820 & 0.3733 & 0.2994 & 0.2437 & 0.3104 & 0.4169 & 0.386 & 0.219 & 0.271 \\

        \bottomrule
    \end{tabular}
\end{table*}

\paragraph{Feature Inputs and Graph Structures.}The results (see Table~\ref{tab:ablations}) show that the global (compressed) variant outperforms the global (multiple) setup across both NLG and CE metrics. We attribute this to the compressed token’s ability to aggregate spatial information across the full feature map, capturing inter-region relationships. In contrast, the multiple-token variant treats spatial locations independently, which limits its ability to model global context, especially under sliding window inference, where each patch has restricted spatial coverage. The local baseline performs comparably to the global variant overall, but shows stronger CE metrics, indicating that region-level features derived from anatomical masks provide valuable semantic cues even without global features. Combining global and local representations yields a modest improvement across all metrics.

\begin{figure}[b]
\centering
\scriptsize
\includegraphics[width=\columnwidth]{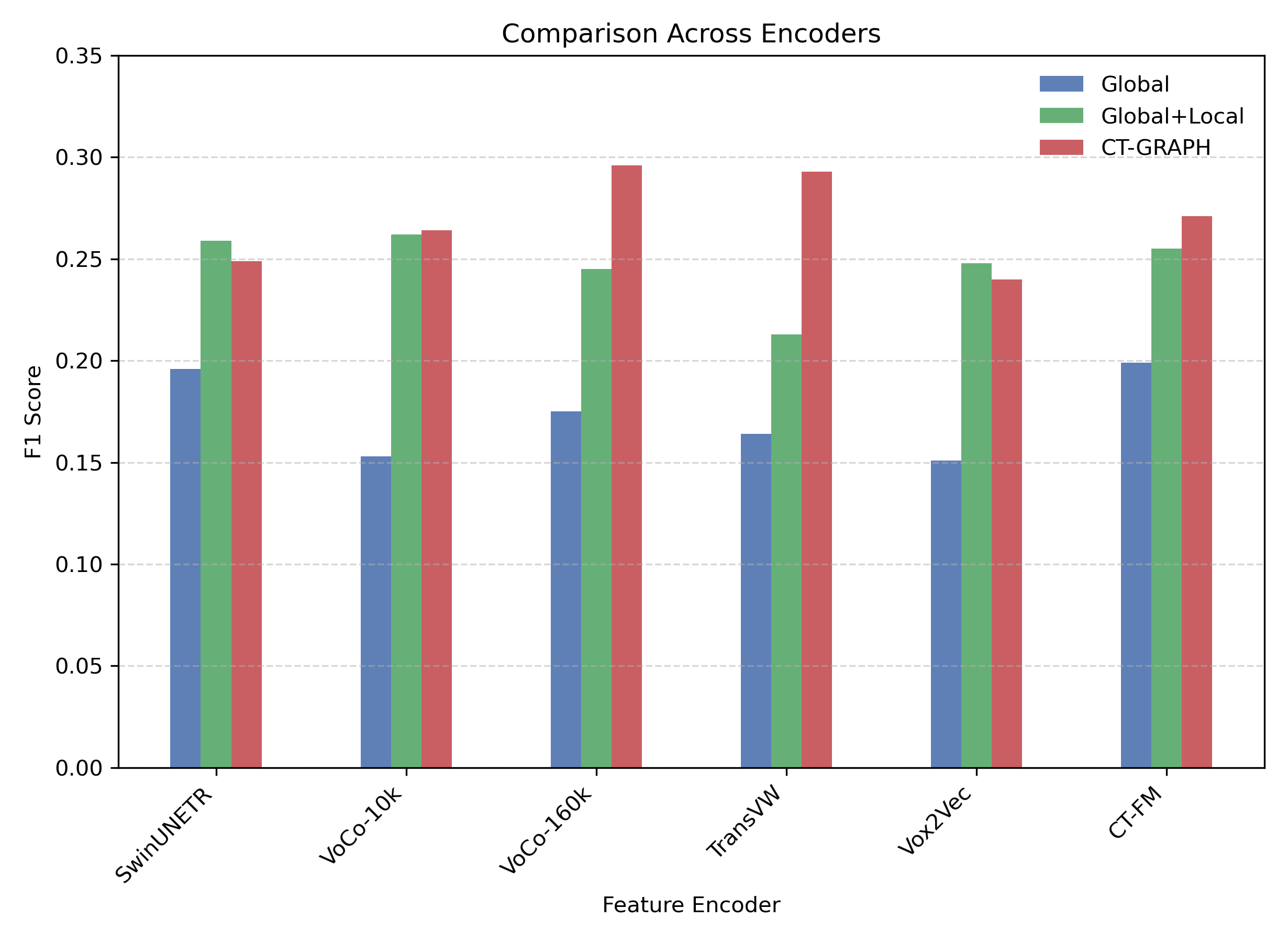}
\caption{F1 score comparison of feature input configurations for across different encoders. } \label{encoder_fig}
\end{figure}

Among the graph-based models, CT-GRAPH achieves the highest overall performance, particularly improving CE F1 from 0.245 to 0.296. This highlights the benefit of explicitly modeling anatomical relationships through hierarchical structure and attention-based message passing. Removing the intermediate hierarchy in the single-level graph leads to slightly lower scores, suggesting that coarse-level nodes contribute meaningfully to performance. The random graph baseline performs substantially worse, reinforcing the importance of anatomically informed graph topology over arbitrary structure.
\paragraph{Effect of Feature Encoders.} To evaluate the generalizability of our method across different pretrained encoders, we compare three input configurations: Global only, Global + Local, and our full model CT-GRAPH across six representative backbones (see Fig. \ref{encoder_fig}). In all cases, the Global + Local setup outperforms the Global-only baseline, confirming that region-level features derived from anatomical masks provide complementary information to global context.

While the magnitude of improvement varies, CT-GRAPH generally matches or exceeds the performance of the unstructured baselines. Particularly strong gains are observed for VoCo-160k and TransVW, while improvements for CT-FM are more modest (e.g., +0.015 F1). Notably, TransVW achieves performance comparable to VoCo-160k despite being pretrained on fewer than 1,000 scans, suggesting that pretraining scale alone does not dictate downstream effectiveness. These differences likely reflect how well encoder features align with anatomical semantics, rather than architectural or dataset size factors alone. Lastly, while NLG scores vary only slightly across encoders, clinical entity metrics show much larger variance indicating that the choice of visual encoder has a greater impact on factual content preservation than on surface-level fluency. Among all feature encoders, VoCo-160k achieves the highest overall performance across metrics, ranking first in every score except precision highlighting its strength as a general-purpose encoder (see Table \ref{tab:feat_evaluation_rg}).

\subsection{Baseline comparisons}

We evaluate the following three competitive baselines for report generation.

\textbf{CT2Rep \cite{hamamci2024ct2rep}.}  
We follow the original training and adapt the preprocessing pipeline to ours as closely as possible, with the exception that we retain the native input resolution of 480$\times$480$\times$240. The model is trained for 20 epochs, consistent with the original setup.

\textbf{CT2Rep w/ LLaMA.}  
To better match our architecture’s language decoding setup, we replace the original decoder in CT2Rep with a pretrained LLaMA2-7B model. The output features from the CTViT encoder are first downsampled to a spatial resolution of 4$\times$4$\times$2 and then passed through a transformer encoder. All 32 tokens are fed into the language model and the same input prompt as for our method is utilized. This variant is trained for 6 epochs to align with our method’s training schedule.

\textbf{Reg2RG \cite{chen2024large}.}  
We adapt the official implementation to use TotalSegmentator region masks. Since masks for the breasts are unavailable, this region is excluded. Due to its low prevalence in the dataset, this omission is expected to have minimal impact on the results. Given their anatomical proximity, we utilize the lung masks for the pleura region. This decision is based on findings in \cite{chen2024large}, where these two regions were not reliably distinguishable. Input volumes are resampled to 256$\times$256$\times$128, which both matches the resolution used for the coronal and sagittal planes during RadFM pretraining and avoids the high memory costs associated with upsampling to 512$\times$512$\times$256.

\paragraph{Overall performance.} 
Table~\ref{tab:final_eval} shows that CT-GRAPH achieves the strongest overall performance across clinical entity (CE) metrics, with the highest recall and F1 score. In terms of natural language generation (NLG), it performs competitively, matching or slightly exceeding the best models across most metrics. These results highlight the effectiveness of CT-GRAPH’s fine-grained, hierarchical feature design in producing clinically informative and linguistically coherent reports.

Both CT2Rep variants yield comparable NLG scores, suggesting that with sufficient training, even the standard decoder can approach the generation quality of a LLM. However, CT2Rep w/ LLaMA shows clear improvements in entity-level accuracy, though it still lags behind CT-GRAPH. Reg2RG achieves similar F1 to CT2Rep w/ LLaMA but slightly underperforms on NLG metrics, possibly due to limitations in the alignment of the provided region-wise reports and TotalSegmentator masks. Overall, while LLMs and region-aware models provide partial gains, neither alone achieves the consistency or balance offered by CT-GRAPH’s structured, multi-scale approach.

\begin{table*}[t]
    \centering
    \scriptsize
    \caption{Performance comparison of our final method (CT-GRAPH + VoCo-160k features) with CT2Rep and Reg2RG.}
    \label{tab:final_eval}
    \renewcommand{\arraystretch}{1.1}
    \setlength{\tabcolsep}{4pt}
    \begin{tabular}{lcccccc|ccc}
        \toprule
        \multirow{2}{*}{\textbf{Method}} &
        \multicolumn{6}{c|}{\textbf{NLG Metrics}} &
        \multicolumn{3}{c}{\textbf{CE Metrics}} \\
        \cmidrule(lr){2-7} \cmidrule(lr){8-10}
        & BLEU-1 & BLEU-2 & BLEU-3 & BLEU-4 & ROUGE-L & METEOR & Precision & Recall & F1 \\
        \midrule
        CT2Rep \cite{hamamci2024ct2rep} & 0.4872 & 0.3712 & 0.2959 & 0.2390 & 0.2992 & 0.4132 & 0.230 & 0.140 & 0.141\\
        CT2Rep (w/ LLaMA)          & \textbf{0.4875}      & 0.3752  & 0.2989  & 0.2426      & 0.3027      & \textbf{0.4212}      &  0.317   & 0.172  & 0.214   \\
        Reg2RG \cite{chen2024large}                 & 0.4728  & 0.3681  & 0.2975  & 0.2437  & 0.2725  & 0.4005 & \textbf{0.457} & 0.156  & 0.217 \\
        CT-GRAPH (ours)             & 0.4850 & \textbf{0.3765} & \textbf{0.3024} & \textbf{0.2467} & \textbf{0.3126} & 0.4205 & 0.396 & \textbf{0.248} & \textbf{0.296} \\
        \bottomrule
    \end{tabular}
\end{table*}

\begin{table}[b]
\centering
\scriptsize
\caption{Per–pathology F1 comparison and class prevalence (Prev). For CT2Rep, the LLaMA variant is evaluated.}
\label{tab:pathology_comparison}
\renewcommand{\arraystretch}{1.1}
\setlength{\tabcolsep}{4pt}
\begin{tabular}{l|ccc|c}
\toprule
\textbf{Pathology} & \textbf{CT2Rep} & \textbf{Reg2RG} & \textbf{CT-GRAPH} & \textbf{Prev.} \\
\midrule
Medical material                    & 0.094 & 0.013 & \textbf{0.284} & 10.5\% \\
Arterial wall calcification         & 0.446 & 0.367 & \textbf{0.604} & 28.9\% \\
Cardiomegaly                        & 0.131 & \textbf{0.390} & 0.208 & 10.9\% \\
Pericardial effusion                & 0.014 & 0.018 & \textbf{0.136} & 7.3\% \\
Coronary artery wall calcification  & 0.411 & 0.378 & \textbf{0.543} & 25.8\% \\
Hiatal hernia                       & 0.138 & 0.075 & \textbf{0.160} & 14.5\% \\
Lymphadenopathy                     & 0.212 & 0.038 & \textbf{0.246} & 26.4\% \\
Emphysema                           & 0.271 & 0.201 & \textbf{0.314} & 20.3\% \\
Atelectasis                         & 0.275 & 0.245 & \textbf{0.279} & 24.2\% \\
Lung nodule                         & 0.465 & 0.383 & \textbf{0.480} & 46.8\% \\
Lung opacity                        & 0.326 & \textbf{0.534} & 0.449 & 40.2\% \\
Pulmonary fibrotic sequela          & 0.188 & 0.125 & \textbf{0.223} & 28.4\% \\
Pleural effusion                    & 0.400 & \textbf{0.574} & 0.546 & 12.6\% \\
Mosaic attenuation pattern          & 0.058 & 0.118 & \textbf{0.133} & 8.4\% \\
Peribronchial thickening            & 0.062 & 0.022 & \textbf{0.082} & 11.6\% \\
Consolidation                       & 0.185 & 0.253 & \textbf{0.382} & 19.8\% \\
Bronchiectasis                      & 0.071 & 0.032 & \textbf{0.109} & 11.2\% \\
Interlobular septal thickening      & 0.104 & 0.138 & \textbf{0.150} & 8.2\% \\
\bottomrule
\end{tabular}
\end{table}

\paragraph{Per-pathology performance.} Table~\ref{tab:pathology_comparison} shows that CT-GRAPH achieves the highest F1 score on 15 of 18 pathologies, outperforming both global (CT2Rep) and region-based (Reg2RG) baselines. The model performs well across a range of conditions, including focal findings like \textit{lung nodule}  and \textit{consolidation}, as well as more diffuse or structurally complex ones such as \textit{emphysema} and \textit{pulmonary fibrotic sequela}. These gains reflect the advantage of modeling fine-grained anatomical units (e.g., lung lobes) and aggregating them hierarchically, allowing CT-GRAPH to reason across spatial scales. Notably, it also outperforms Reg2RG on other localized abnormalities, such as \textit{arterial wall calcification} (0.604 vs.\ 0.367), where finer spatial granularity enables more precise feature modeling than Reg2RG’s coarser, whole-organ segmentation. In contrast, CT2Rep, which relies solely on global volumetric features, struggles on small or spatially concentrated findings and fails to lead on any pathology class, highlighting the limitations of global features for fine-grained clinical detail.

\section{Conclusion}
\label{sec:conclusion}

In this work, we presented CT-GRAPH, a novel approach for CT report generation that effectively leverages pretrained 3D feature encoders and anatomically structured graph representations. We conducted an extensive evaluation of how pretrained features could be effectively utilized, focusing on both global and organ-specific features to maintain fine-grained spatial context. 
We have shown the effect of multi-layer fusions for the recognition of volume-level pathologies as well as the effect of different pooling level strategies. We have further demonstrated how these features can be aggregated for report generation purposes, highlighting the limitations of relying solely on either global or local features for report generation. In contrast, our proposed hierarchical graph attention network effectively integrates multi-level anatomical information, improving the model’s ability to produce accurate, anatomy-aware reports. Through extensive experimentation, we showed that our method not only outperformed the baseline setups but also existing concurrent work, achieving significant improvements, particularly in clinical efficacy metrics.

{
    \small
    \bibliographystyle{ieeenat_fullname}
    \bibliography{main}
}


\end{document}